\documentclass{ar-1col-S2O}
\usepackage[numbers]{natbib}
\usepackage{url}
\usepackage{enumitem}
\usepackage{lmodern}
\usepackage{comment}
\usepackage[hidelinks]{hyperref}
\usepackage{tcolorbox}
\usepackage{ifthen}
\usepackage{bm}

\newcommand{\showmarginnote}{1}

\usepackage{amsmath}
\usepackage{amsfonts}
\jname{Preprint to appear in the 2026 Annual Review of Control, Robotics, and Autonomous Systems, Vol. 9}
\jyear{2026}

\definecolor{international_orange}{RGB}{240, 74, 0}

\newcommand{\todo}[1]{\textcolor{red}{\textbf{[TODO: #1]}}}
\newcommand{\Symptom}{\textcolor{NavyBlue}{\textbf{Symptom: }}}

\newcommand\blfootnote[1]{%
  \begingroup
  \renewcommand\thefootnote{}\footnote{#1}%
  \addtocounter{footnote}{-1}%
  \endgroup
}

\usepackage{tikz}
\usetikzlibrary{arrows.meta, positioning, shadows.blur}
\begin{document}

\markboth{Aljalbout et al.}{The Reality Gap in Robotics}

\title{The Reality Gap in Robotics:
Challenges, Solutions, and Best Practices}

\author{Elie Aljalbout$^1$, 
Jiaxu Xing$^1$, 
Angel Romero$^1$,  
Iretiayo Akinola$^2$,
Caelan Reed Garrett$^2$,\\
Eric Heiden$^2$,
Abhishek Gupta$^4$, 
Tucker Hermans$^{2,5}$, 
Yashraj Narang$^2$,
Dieter Fox$^4$,
Davide Scaramuzza$^1$, 
Fabio Ramos$^{2,3}$
\affil{$^1$Robotics and Perception Group, University of Zurich, Zurich, Switzerland, 8050
}
\affil{$^2$NVIDIA, Seattle, USA, 98105}
\affil{$^3$The University of Sydney, Sydney, Australia, 2006}
\affil{$^4$University of Washington, Seattle, USA, 98195}
\affil{$^5$University of Utah, Salt Lake City, USA, 84112}
}


\begin{abstract}
Machine learning has facilitated significant advancements across various robotics domains, including navigation, locomotion, and manipulation.
Many such achievements have been driven by the extensive use of simulation as a critical tool for training and testing robotic systems prior to their deployment in real-world environments.
However, simulations consist of abstractions and approximations that inevitably introduce discrepancies between simulated and real environments, known as the reality gap. 
These discrepancies significantly hinder the successful transfer of systems
from simulation to the real world. 
Closing this gap 
remains one of the most pressing challenges in robotics.
Recent advances in sim-to-real transfer have demonstrated promising results across various platforms, including locomotion, navigation, and manipulation. 
By leveraging techniques such as domain randomization, real-to-sim transfer, state and action abstractions, and sim-real co-training, many works have overcome the reality gap. 
However, challenges persist, and a deeper understanding of the reality gap's root causes and solutions is necessary.
In this survey, we present a comprehensive overview of the sim-to-real landscape, highlighting the causes, solutions, and evaluation metrics for the reality gap and sim-to-real transfer. 
\end{abstract}

\maketitle


\section{Introduction}

Simulation holds great potential for robot learning, benchmarking and large-scale data collection in robotics due to its scalability, safety, and efficiency. Robots can be safely trained in simulation before being deployed to the real world. Algorithms can be compared against each other in simulation over multiple simulated scenarios. Through simulation, massive amounts of data can be collected from robots performing complex tasks at a fraction of the cost compared to the real world.\blfootnote{Website: \url{https://robotics-reality-gap.github.io/}}
However, the gap between simulated and real-world environments often stands in the way of fully leveraging its potential.
Bridging the gap between simulation and the real world has become one of the most critical and long-standing challenges in robotics.
Overcoming this challenge holds the potential to accelerate progress in robotics similar to what has been achieved in other computational fields.
For example,
sustained progress observed in the natural language processing and computer vision communities would not have been possible without significant effort in the development of large training and benchmarking datasets~\cite{kirillov2023segment,achiam2023gpt}. 
In contrast, robotics still lags behind in these aspects. 
One reason for this is that static test datasets cannot reflect the complexity of perceiving and acting in the real world. 
Instead, most robotics problems require training and benchmarking in interactive environments. 
While training in the real world has the benefit of making sure that systems are trained and tested in realistic settings, complex real-world tasks are difficult to scale and replicate with sufficient reproducibility.
In addition, data collection in the real world is bottlenecked by multiple factors such as hardware cost, human dependence, and difficulty of automation.
Simulation offers an affordable alternative for these challenges, making it an invaluable tool for data collection, benchmarking, and building various components of typical robotic pipelines, such as perception, planning, learning, and control. 
Namely, for benchmarking, simulation ensures the repeatability of experiments, which is impossible in real-world robotics systems due to their stochastic nature.
For data generation, simulation provides an efficient approach that can leverage multiple robots operating faster than in real-time and in a parallelized manner.
\begin{figure}
    \centering
    \includegraphics[width=\linewidth]{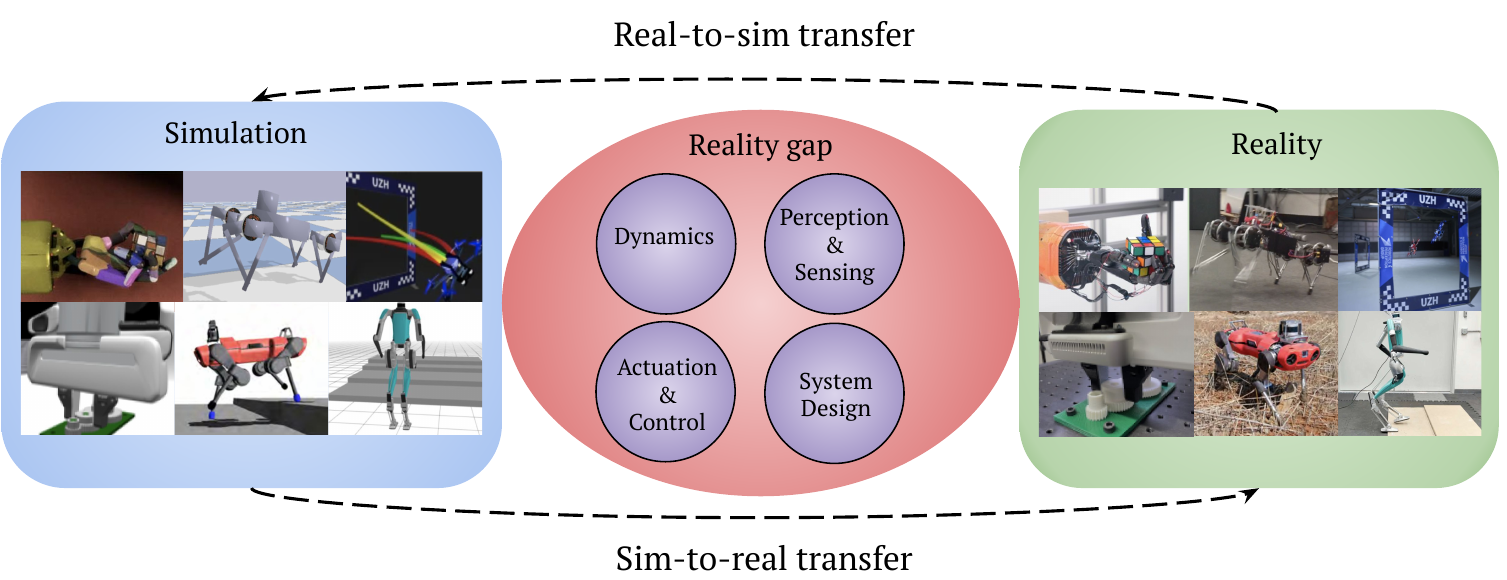}
    \caption{The reality gap is the difference between a simulated and real environment in aspects related to dynamics, perception \& sensing, actuation \& control, and system design. 
    By carefully designing these modules, the gap can be reduced to a reasonable size.
    Sim-to-real as well as real-to-sim transfer require methods that can carefully overcome the remaining reality gap.
    Figures are adapted from successful sim-to-real applications in various robotics domains~\cite{akkaya2019solving, tan2018sim,Tang-RSS-23, lee2020learning, radosavovic2024real,kaufmann2023champion}.}
    \label{fig:transer_overview}
\end{figure}

Simulation attempts to replicate physical reality with mathematical abstractions, models, and approximations.
This means that there are no perfect simulators, and therefore there is always a difference from the real world, which we call \emph{reality gap}.
While it is a common misconception to talk about this gap as a single element, the reality gap consists of a large number of sub-gaps resulting from the simulation's failure to replicate various physical real-world mechanisms and phenomena accurately.
As a consequence, transferring any sort of behavior observed in simulation to a real-world environment can be extremely challenging.
For instance, transferring control policies designed or learned in simulation is not trivial~\cite{muratore2019assessing}.
Due to the differences between the simulated and the real environments, policies obtained in simulation could achieve great performance in simulation, simply by abusing modeling inaccuracies and simulator-specific corner cases.
Hence, the successful transfer of such policies is not guaranteed and can even be dangerous to the robot, its surroundings, and any human in its proximity. 

Despite these challenges, recent progress in robot learning has shown great promise in sim-to-real transfer, where control policies learned in a simulation are transferred to a similar real-world environment, and multiple techniques have been proposed to overcome the reality gap for different robotic platforms~\cite{rusu2017sim, sadeghi2017cad2rl, tobin2017domain, nagabandi2018learning, akkaya2019solving, andrychowicz2020learning, lee2020learning, Tang-RSS-23, kaufmann2023champion, zhuang2023robot}.
Besides improvements in physics engines and rendering quality, simulation technology progressed to meet the demands of the robot-learning community by following the deep learning trend of massive parallelization and large-scale data collection.
Modern simulators can leverage GPU parallelization to simulate thousands of robots simultaneously~\cite{mittal2023orbit,makoviychuk2021isaac,liang2018gpu,todorov2012mujoco}.
This ability to efficiently parallelize simulations has led to significant breakthroughs in the field, particularly in locomotion~\cite{lee2020learning, rudin2022learning}, agile flight~\cite{song2023reaching,kaufmann2023champion}, and manipulation~\cite{akkaya2019solving, andrychowicz2020learning, zhanglearning, fu2023deep}.
In locomotion, highly parallelized simulation enabled learning very robust quadruped locomotion across challenging terrains~\cite{lee2020learning} and, more recently bipeds control~\cite{Grandia-RSS-24, radosavovic2024real}.
Furthermore, sim-to-real transfer has been used for dexterous manipulation as well as other various single-arm~\cite{10522877,Tang-RSS-23} and dual-arm tasks~\cite{alles2022learning}, ranging from simple 3D reaching to more complex and contact-rich tasks such as table-top rearrangement~\cite{10522877} and assembly~\cite{alles2022learning,Tang-RSS-23}.
Additionally, sim-to-real transfer played a key role in learning agile quadrotor control policies that outperformed human champions in drone racing~\cite{song2023reaching, kaufmann2023champion}.

However, sim-to-real still faces many challenges, such as photorealistic rendering (more generally, sensor modeling) and simulation of complex dynamics such as contacts, deformations, and material variations.
To further progress sim-to-real transfer and understand its limitations, a common and structured understanding of the problem is necessary.
Ideally, such an understanding should go all the way from the roots of the problem to the existing and needed solutions to overcome them.

In this survey, we provide a comprehensive overview of the sim-to-real landscape for robot learning.
We dissect the problem into atomic components, identify the sources of the reality gap and the symptoms they cause, and provide metrics and solutions to understand these problems and alleviate them in practice.
Our objective is to boost the understanding of the problem by providing a guide for researchers and practitioners.
We first introduce the problem, its notation and its theory in section~\ref{sec:preliminaries}.
In section~\ref{sec:gaps}, we identify and exhaustively list the different causes of the reality gap.
We then survey existing solutions and metrics for sim-to-real transfer in section~\ref{sec:existing} and section~\ref{sec:evaluation}, respectively.
Finally, in section~\ref{sec:open}, we discuss open problems and an outlook for future research on this topic.

\section{Preliminaries} \label{sec:preliminaries}
In this section, we formalize the concepts and definitions required to understand and characterize the reality gap as well as sim-to-real transfer.

\subsection{States, Observations, and Actions}

We first introduce a general mathematical model for robot systems that receive observations of the world using their sensors and make decisions that maximize a particular objective.
We formulate this decision-making problem as a Partially Observable Markov Decision Process (POMDP) \footnote{When the sensors provide the full state without noise (e.g. in simulation with privileged information) observations are equivalent to states and the formulation collapses to a Markov Decision Process (MDP).}.
A POMDP is defined as the tuple $\mathcal{M} = (\mathcal{S}, \mathcal{A}, \mathcal{T}, \mathcal{R}, \mathcal{Z}, \mathcal{O}, \gamma)$, where 
$\mathcal{S}\subseteq\mathbb{R}^n$ denotes the state space encompassing all feasible configurations of the robot and its environment.
$\mathcal{A}\subseteq\mathbb{R}^m$ is the action space comprising all control commands the robot can execute;
$\mathcal{T}: \mathcal{S} \times \mathcal{A} \rightarrow \mathcal{S}$ denotes the transition dynamics, describing how the state evolves given the current state $\mathbf{s}_t$ at time $t$ and a given action $\mathbf{a}_t$; 
$\mathcal{R}:\mathcal{S}\times\mathcal{A}\rightarrow\mathbb{R}$ is the reward function encoding the task objective;
$\mathcal{Z}\subseteq\mathbb{R}^h$ is the observation space, and
$\mathcal{O}:\mathcal{S}\!\rightarrow\!\mathcal{Z}$ is the sensor (observation) model that maps the latent state to an observation $\mathbf{z}_t$; finally,
$\gamma\in[0,1)$ is the discount factor, representing the relative importance of future rewards with respect to immediate rewards.
The agent maintains a belief $\mathbf{b}_t\in \mathcal{B}(\mathcal{S})$, a probability distribution over states, updated from the history of past actions and observations.
Its goal is to learn a policy
$\pi:\mathcal{B}(\mathcal{S})\!\rightarrow\!\mathcal{A}$ that maximizes the expected discounted return
\begin{equation}
\pi^{*}= \arg\max_{\pi}\;
\mathbb{E}\!\left[\;\sum_{t=0}^{\infty}\gamma^{t}\,
\mathcal{R}\!\bigl(\mathbf{s}_t,\pi(\mathbf{b}_t)\bigr)\right],
\end{equation}
where $\mathcal{R}(\mathbf{s}_t, \pi(\mathbf{b}_t))$ is the reward the agent receives when in belief $\mathbf{b}_t$ acting under the policy $\pi$.
In practice, sim-to-real work often represents $\mathbf{b}_t$ by a compact feature vector extracted directly from raw observations such as camera images and proprioceptive
signals.

\subsection{Simulation}

A simulation is a computational approximation of the real world. 
If we denote the real world dynamics perfectly capturing the behavior of the real world system as $\mathcal{T}_r(\mathbf{s}_{t+1} | \mathbf{s}_t, \mathbf{a}_t)$, 
and the real observation model
$\mathcal{O}_r(\mathbf{z}_t | \mathbf{s}_t)$, a simulation approximates the real world dynamics and observation model by employing physical models derived from first principles, numerical integration, and approximations to reduce complexity and computational cost, such that $\mathcal{T}_\mathbf{s}(\mathbf{s}_{t+1}|\mathbf{s}_t, \mathbf{a}_t) \approx \mathcal{T}_r(\mathbf{s}_{t+1}|\mathbf{s}_t, \mathbf{a}_t)$ and $\mathcal{O}_\mathbf{s}(\mathbf{z}_t | \mathbf{s}_t)\approx\mathcal{O}_r(\mathbf{z}_t | \mathbf{s}_t),$ $
\forall\,\mathbf{s}_t\in\mathcal{S},\,\mathbf{a}_t\in\mathcal{A}$.
A perfect simulator aims to minimize the discrepancy between the simulated and real world dynamics as well as the geometric structure and the simulated observations. 
However, perfect fidelity is impractical and computationally infeasible in realistic robotic scenarios due to the inherent complexity of physical effects (friction, noise, sensor latency, etc).

\subsection{Reality Gap}
\label{sec:realitygap}

We distinguish between the \emph{reality gap} characterizing the difference between the simulated and the real environments, and the \emph{performance gap}, characterizing the difference in performance in simulation and the real world for a given policy.

The \emph{reality gap} is the gap between the simulated POMDP $\mathcal{M}_\text{s}$ and the real one $\mathcal{M}_r$.
It is mainly composed of the dynamics and perception gaps,
\begin{equation}
\begin{aligned}
     G_{\text{dyn}}(\mathcal{M}_\text{s}, \mathcal{M}_\text{r}) &=\mathbb{E}_{(\mathbf{s},\mathbf{a}) \sim \mathcal{M}_r} \left[ D\left( T_{\text{sim}}(\cdot \mid \mathbf{s}, \mathbf{a}) \,\|\, T_{\text{real}}(\cdot \mid \mathbf{s}, \mathbf{a}) \right) \right] \\
     G_{\text{perc}}(\mathcal{M}_\text{s}, \mathcal{M}_\text{r}) &=\mathbb{E}_{\mathbf{s} \sim \mathcal{S}_\text{r}} \left[ D\left( O_{\text{sim}}(\cdot \mid \mathbf{s}) \,\|\, O_{\text{real}}(\cdot \mid \mathbf{s}) \right) \right],
\end{aligned}
\end{equation}
where $D$ is a measure of divergence or dissimilarity, $T_{\text{real}}$, $T_{\text{sim}}$ and $O_{\text{real}}$, $O_{\text{sim}}$ refer to transition dynamics and observation models with subscripts $\text{real}$ and $\text{sim}$ indicating the real and simulated environments. 

The performance gap is the difference in performance of a given policy when executed in simulation versus in the real world.
Let
$\pi : \mathcal{B}(\mathcal{S}) \!\rightarrow\! \mathcal{A}$
denote a policy trained in an environment
$\mathcal{M}$, where $\mathcal{B}(\mathcal{S})$ is the belief space.
Starting from an initial state $\mathbf{s}_0$, the closed-loop trajectory
$\tau = (\mathbf{s}_{0},\mathbf{z}_{0},\mathbf{a}_{0},\mathbf{s}_{1},\mathbf{z}_{1},\mathbf{a}_{1},\dots,\mathbf{s}_{T},\mathbf{z}_{T})$
evolves under transition dynamics $\mathcal{T}$ and observation model
$\mathcal{O}$.
The expected discounted return of $\pi$ in environment
$\mathcal{M}$ is

\begin{equation}
J_{\mathcal{M}}(\pi) = \mathbb{E}_{\tau \sim p(\tau|\pi, \mathcal{M})}\left[\sum_{t=0}^{T}\gamma^{t}R(\mathbf{s}_t,\pi(\mathbf{b}_t))\right].
\end{equation}
The performance gap $G_{\text{perf}}(\mathcal{M}_\text{s}, \mathcal{M}_\text{r},\pi)$ for the policy $\pi$ is formally defined as the absolute difference between its expected performance in the simulated and real environments:
\begin{equation}
    G_{\text{perf}}(\mathcal{M}_\text{s}, \mathcal{M}_\text{r},\pi) = \left| J_{\mathcal{M}_\text{s}}(\pi) - J_{\mathcal{M}_\text{r}}(\pi) \right|.
\end{equation}
Minimizing $G_{\text{perf}}(\mathcal{M}_\text{s}, \mathcal{M}_\text{r},\pi)$ ensures effective transfer and robust performance from simulation to real-world scenarios.
It is important to note that exact replication of real-world dynamics (i.e., elimination of $\mathcal{G}_{\text{dyn}}$) and observation models (i.e., elimination of $\mathcal{G}_{\text{obs}}$) in a simulator is not practically achievable nor required to achieve successful sim-to-real gap minimization.
Successful transfer can still occur if the policy is robust against differences between the environments. 
Therefore, the objective of sim-to-real transfer is
\begin{align}
    \pi^* = \min_{\pi} G_{\text{perf}}(\mathcal{M}_\text{s}, \mathcal{M}_\text{r},\pi).
\end{align}
\section{Sources of Reality Gap}  
\label{sec:gaps}

In this section we elaborate on the differences in dynamics (Section~\ref{subsec:dynamics}), perception (Section~\ref{subsec:perception}), actuation (Section~\ref{subsec:actuation}), and system design (Section~\ref{subsec:system}), all contributing to the sim-to-real gap.

\begin{figure}
    \centering
    \input{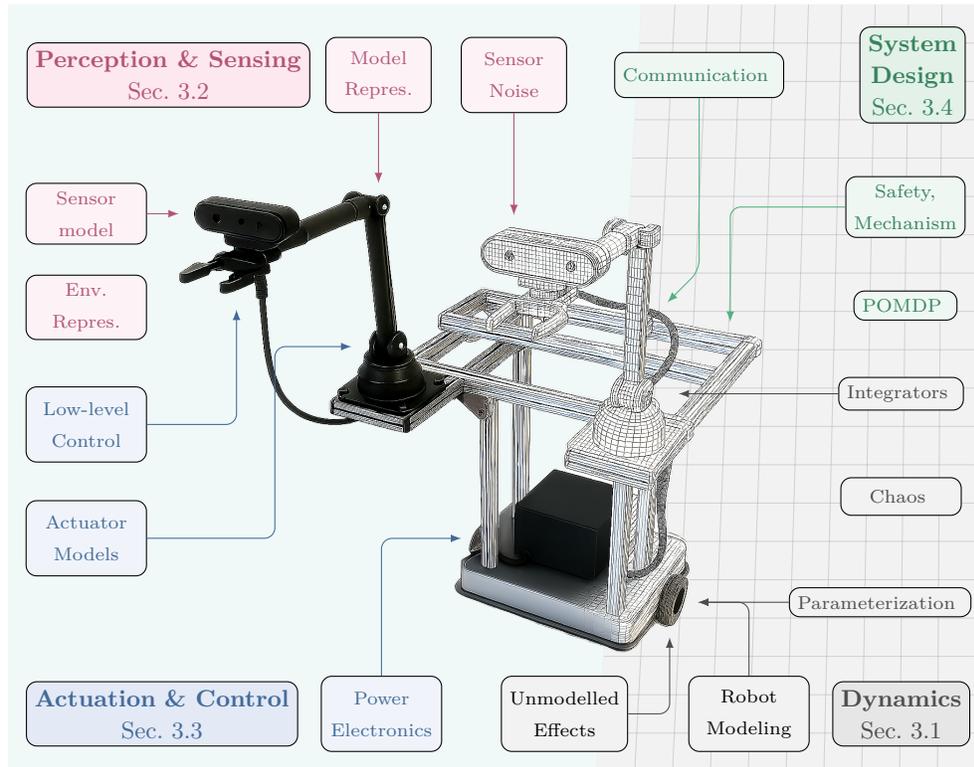}
    \caption{Illustration of the different sources of the reality gap in the \textcolor{Emerald!70}{real world} (left) and \textcolor{black!40}{simulation} (right).}
    \label{fig:gap_list}
\end{figure}

\subsection{Dynamics}
\label{subsec:dynamics}
One of the most important causes of the reality gap is the dynamics of the system.
The \emph{dynamics gap} $G_{\text{dyn}}(\mathcal{M}_s, \mathcal{M}_r)$ is defined by sim-to-real discrepancies in the transition model $\mathcal{T}$ in the POMDP formulation.
When creating a simulator, many decisions and simplifications are made about what to model and what to leave out, how to model dynamics, what parameters best approximate the real world, and how to represent continuous phenomena in discrete computations.
These decisions and assumptions can lead to a plethora of dynamics gaps.
A policy trained in a simulation with inaccurate dynamics may learn to exploit these inaccuracies, leading to a potential performance drop in the real world.
In the following, we list the most important sources.
\subsubsection{Modeling}
Simulators attempt to replicate real-world dynamics with models.
The models represent various aspects of physics, such as rigid-body dynamics, batteries, chaotic nature, and stochasticity. 
Each of these components can contribute to the reality gap.
\vspace{0.5em}
\begin{itemize}[leftmargin=*, nosep]
      \item \textbf{Rigid body dynamics.}
        Most robot simulators assume perfectly rigid bodies and joints, but objects can be deformable, and real robots can be flexible, compliant or have uneven joints.
        For example, real robot links and frames can bend or vibrate under load, whereas simulators usually treat them as rigid bodies.
        Similarly, simulated joints are often modeled as ideal, but they suffer from damping, internal springs, backlash, etc.
        \ifthenelse{\equal{\showmarginnote}{1}}{
        \begin{marginnote}
            \footnotesize \Symptom
            A policy trained with rigid-body assumptions may lead to unexpected behaviors when dealing with  real, deformable objects.
        \end{marginnote}
        }
        {}
        This mismatch is a fundamental source of errors in dynamics modeling.  
      
    \item \textbf{Chaotic nature.}
        Some real-world phenomena exhibit chaotic behavior, characterized by a sensitive dependence on initial conditions, making them inherently non-reproducible even with perfect models.
        These are fundamentally difficult to capture with a dynamic model in a simulation.
        For example, atmospheric turbulence, complex flow patterns and pressure waves in fluid dynamics cannot be fully captured by simplified models.
        \ifthenelse{\equal{\showmarginnote}{1}}{
        \begin{marginnote}
            \footnotesize \Symptom
            Due to chaotic effects, an overfitted policy with slightly different initial conditions might fail in an unpredictable manner.
        \end{marginnote}
        }
        {}

    \item \textbf{Stochasticity.}
        The real world contains numerous sources of stochastic dynamics that are fundamentally difficult or impossible to represent in simulation.
        For instance, ground-based robots (wheeled or legged) encounter stochastic surface and friction variations and debris that create unpredictable disturbances.
        Simulators typically either ignore these effects entirely or model them as simplified Gaussian noise, failing to capture the complex spatiotemporal correlations and non-linear coupling effects present in reality. 

    \item \textbf{Battery.}
        This is particularly relevant for \emph{mobile robots}. Even if the actuator model is accurate, the energy source that feeds it can introduce inaccuracies.
        Since motor torque scales directly with the voltage that the battery provides, any drop in battery voltage immediately reduces the maximum joint torque the hardware can deliver. 
        Hence, rapid joint accelerations can experience a transient torque deficit that is absent in simulation.
        Batteries are rarely modeled because their behavior is very nonlinear and history-dependent; terminal voltage drops with instantaneous load and varies with cell temperature and age. 
        \ifthenelse{\equal{\showmarginnote}{1}}{
        \begin{marginnote}
       \footnotesize \Symptom
       Unmodeled battery dynamics (e.g., voltage drop with load or age) may cause the policy to produce torques that are less than required to achieve the desired real-world motion.\end{marginnote}Neglecting these effects in simulation leaves controllers unaware of the reduced control authority they encounter on real hardware, especially when operating at the limit of the robot's performance.
       }
       {}

\item \textbf{Contact Dynamics.}
A key contributor to the dynamics gap in robotics is the inaccurate modeling of physical contact.
Real-world contact interactions are inherently complex and nonlinear. 
Materials deform under pressure, friction varies with relative velocity, and contact states can alternate between sticking, slipping, and separation.
To maintain computational efficiency, simulators typically rely on simplified models such as point contacts, linearized friction cones, or compliant spring-damper systems.
While these approximations enable faster simulation, they often fail to capture critical physical behaviors. 
This can result in non-physical artifacts such as spurious internal forces, unstable grasps, or unrealistic motion patterns.
Such mismatches are particularly problematic in contact-rich tasks like robotic manipulation, where precise interaction modeling is crucial for successful control and planning.
\end{itemize}
\ifthenelse{\equal{\showmarginnote}{1}}{
\begin{marginnote}
\footnotesize \Symptom
A policy may fail to grasp objects because its learned model is based on simplistic contact assumptions, leading to unexpected slipping or instability.
\end{marginnote}
}
{}
\subsubsection{Parameterization}
A fundamental source of the dynamics gap arises from the incorrect parameterization of the physical model.
While simulators account for various physical parameters such as friction, aerodynamics, mass, inertia, etc., assigning accurate values to these parameters can be challenging.
Some physical properties can be difficult to measure precisely and can be subject to changes over time.
\ifthenelse{\equal{\showmarginnote}{1}}{
\begin{marginnote}
    \footnotesize \Symptom
     A policy may overshoot or undershoot because its learned dynamics are based on incorrect parameters.
\end{marginnote}
}
{}
\subsubsection{Numerical Integrators}
The choice of numerical integration methods for unrolling differential equations significantly impacts the reality gap.
Simulators rely on numerical integration schemes to approximate continuous dynamics.
Discretization and the specific integration method employed (e.g., Euler, Runge-Kutta, Quadrature, etc) can introduce discrepancies between the simulated and the real system.\ifthenelse{\equal{\showmarginnote}{1}}{\begin{marginnote}
    \footnotesize \Symptom
    Performance may degrade over long tasks as small integration errors accumulate.
\end{marginnote}}{}Additionally, there is a trade-off between fidelity and computation time.
Increasing the numerical integrator's accuracy by using smaller time steps or higher-order methods, generally leads to more accurate simulations at the cost of increased computational time.

\subsubsection{Human-Robot Interaction}
Modeling human behavior in simulation presents unique challenges that create substantial reality gaps~\cite{DAmbrosioJSOXLS23}.
Humans exhibit complex, context-dependent, and often irrational behaviors that are difficult to capture with computational models.\ifthenelse{\equal{\showmarginnote}{1}}{\begin{marginnote}
    \footnotesize \Symptom
    A policy could take advantage of wrong assumptions in the simulated human behavior.
\end{marginnote}}{}Therefore, humans are typically simulated as simplified agents with predefined motion patterns, and basic reactive responses to robot actions.
\subsubsection{Unmodeled Effects}
There is a range of physical phenomena that are often overlooked or simplified in simulation, and that can cause significant sim-to-real differences.
These include wear and tear of robot components, which degrade over time due to friction, material fatigue, repetitive usage, etc.\ifthenelse{\equal{\showmarginnote}{1}}{\begin{marginnote}[2em]
    \footnotesize \Symptom
    Performance may drift over time, or the robot may exhibit unexpected vibrations and instability as its physical properties change.
\end{marginnote}}{}These degradations lead to changes in material properties such as stiffness, backlash, vibrations, or texture, which are not typically captured in simulators.
Thermal effects can also add to this category, as temperature fluctuations can affect the performance of motors, sensors, batteries, and other components, potentially leading to a significant change in their behavior.
\subsubsection{Asset Fidelity} 

Simulations represent the environment using digital assets to encode geometry, including primitive shapes, meshes, and Signed Distance Fields (SDFs) and material parameters such as friction coefficients, restitution, density, etc.
However, for computational efficiency reasons, these representations are often simplified and approximated, utilizing clean layouts and low-resolution models that lack the complexity and properties of real-world environments, such as irregular terrain and fine-grained intricate structures.
Similarly, the robot's own geometry is often simplified, omitting important physical details and using simplified shapes for the sake of efficiency, which can lead to unexpected behaviors such as self-collisions or unstable motions in reality.
\ifthenelse{\equal{\showmarginnote}{1}}{
\begin{marginnote}
    \footnotesize \Symptom
    A policy trained with idealized digital assets might act on inaccurate geometric assumptions, leading to collisions with unmodeled effects.
\end{marginnote}
}{}
\subsection{Perception and Sensing}
\label{subsec:perception}
In this section, we discuss the sources of the reality gap originating from perception and sensing.
The \emph{perception gap} is defined by sim-to-real discrepancies in the observation model $O$ in the POMDP formulation.
State-of-the-art simulators with advanced rendering pipelines~\cite{mittal2023orbit, isaacsim2023, dosovitskiy2017carla, shah2018airsim, savva2019habitat} have significantly improved the realism of sensor effects. 
However, they still fall short of capturing the full complexity and variability of the real world, often resulting in discrepancies between simulated and real sensory measurements.
Consequently, policies trained in such simplified settings struggle to generalize during real-world deployment.
In the following, we outline the major sources of these mismatches and their implications for sim-to-real transfer.
\subsubsection{Sensor Model}
\label{sec:sensor_models}
Simulation models are designed to mimic the physical characteristics of real-world sensors, including properties such as field of view, resolution, and noise patterns.
Additionally, they simulate how sensors respond to motion and how they are mounted on the robot since placement significantly influence the resulting sensor data.
\vspace{0.5em}
\begin{itemize}[leftmargin=*, nosep]
    \item \textbf{RGB Cameras.} 
    For RGB cameras, NVIDIA Isaac Sim~\cite{isaacsim2023} uses ray tracing and physically based rendering to generate photorealistic images.
    CARLA~\cite{dosovitskiy2017carla} and AirSim~\cite{shah2018airsim}, both built on Unreal Engine, simulate realistic visuals for navigation.
    These simulators are capable of modeling realistic lighting, shadows, and material interactions. 
    Many platforms optimized for real-time performance, such as Gazebo Classic~\cite{koenig2004design} and early MuJoCo~\cite{todorov2012mujoco}, use simplified OpenGL-based rendering with idealized pinhole camera models and Z-buffer depth computation~\cite{catmull1974subdivision}. 
    This omits real-world effects such as lens flares, chromatic aberration, flying pixels, and rolling shutter distortions, causing learned features to poorly transfer to real-world settings~\cite{tremblay2018training}.
    \ifthenelse{\equal{\showmarginnote}{1}}{
    \begin{marginnote}
        \footnotesize \Symptom
       Sensor-specific artifacts (e.g., lens distortion, LiDAR beam patterns) can lead to massive sim-to-real discrepancies in the distribution of observations.
    \end{marginnote}
    }{}
    \item
    \textbf{Depth Sensors.}
    Many simulators support depth sensor simulation, including stereo cameras, structured light (e.g., Kinect v1), and time-of-flight (e.g., Intel RealSense) systems.
    Platforms such as Isaac Sim~\cite{isaacsim2023}, Habitat-Sim~\cite{savva2019habitat} offer synthetic depth maps derived from rendered 3D geometry. 
    However, these simulations typically assume ideal depth projections and often omit real-world artifacts such as quantization noise, depth shadows, and ambient light interference~\cite{nguyen2012modeling, huang2019deepfusion}. 
    \item
    \textbf{LiDAR Sensors.}
    Simulators such as CARLA simulate LiDAR sensors using raycasting techniques from the sensor origin to the scene geometry.
    While CARLA approximates various real-world effects,  many intricate characteristics remain difficult to reproduce accurately. 
    These include beam divergence patterns, material-dependent reflectivity, angle-of-incidence effects~\cite{laconte2018lidar}, and sensor-specific interference artifacts observed in real devices like Velodyne, Ouster, or Livox. 
\item
\textbf{Other Sensors.}
Simulators also model sensors including joint encoders, IMUs, GPS, and tactile sensors. 
However, they often idealize sensor behavior, neglecting real-world effects such as IMU drift, GPS multipath, or latency. 
These simplifications can introduce reality gaps, especially in tasks relying on sensor fusion or precise state estimation.
\end{itemize}
\subsubsection{Sensor Noise}
Real-world sensor measurements are inherently noisy due to factors such as thermal effects~\cite{susperregi2013use}, quantization errors~\cite{white2008digital}, and environmental interference~\cite{barshan1995noise}.
Crucially, sensor noise is often complex, non-Gaussian, state-dependent, temporally correlated, and influenced by motion, temperature, lighting, and surface properties.
For example, depth sensors exhibit range-dependent uncertainty, structured missing data at depth discontinuities, and surface-dependent noise patterns~\cite{nguyen2012modeling}.
Despite these complexities, many simulators use simple Gaussian noise models with fixed variance. 
\ifthenelse{\equal{\showmarginnote}{1}}{
\begin{marginnote}
    \footnotesize \Symptom
   Policies trained on simple Gaussian noise may overfit and fail under complex, state-dependent, and temporally correlated real-world sensor noise.
\end{marginnote}
}{}
\subsubsection{Environment Representation}
In Section~\ref{subsec:dynamics}, we discussed how asset fidelity affects the accuracy of dynamics simulation. 
A similar issue arises in perception and sensing: using low-resolution assets, overly simplified scene graphs, or generic materials can fail to capture fine-grained perceptual cues such as surface textures, reflectance, and subtle geometry.
Additionally, the lack of High Dynamic Range Image (HDRI) backgrounds may result in unrealistic lighting, while not differentiating between static and dynamic bodies can obscure critical motion and occlusion relationships. 
These limitations significantly degrade perceptual realism, which becomes particularly problematic in tasks requiring fine-grained object recognition or precise physical interaction.
\ifthenelse{\equal{\showmarginnote}{1}}{
\begin{marginnote}
   \footnotesize \Symptom
    The robot may fail to recognize objects or get lost in a real room that looks different from the visually simplistic simulation.
\end{marginnote}
}{}

\subsubsection{Robot Model}
Simulated robots are typically defined by their geometry, kinematics, and dynamics.
These models are often based on CAD files and URDF or USD descriptions.
While accurate in structure, they usually simplify or omit important physical details.
Real-world factors such as manufacturing tolerances, material wear, and mechanical backlash are rarely modeled. 
These discrepancies can introduce reality gaps. 
It can cause self-collisions, unstable motions, or failed task execution~\cite{peng2020learning, arm2024pedipulate}.
\subsubsection{Collision Sensing}
Simulators rely on efficient collision detection algorithms to determine contact events between the robot and its environment, or within the robot itself. 
To achieve real-time performance, they typically use simplified geometric approximations, such as bounding volumes, convex decompositions, or sphere decompositions—instead of high-resolution visual meshes~\cite{coumans2021, todorov2012mujoco}. 
These proxy shapes are evaluated at discrete time steps, which further limits the accuracy of contact modeling, especially in tasks involving fine-grained manipulation or dense contact.
\ifthenelse{\equal{\showmarginnote}{1}}{
\begin{marginnote}
    \footnotesize \Symptom
    When attempting dexterous manipulation, the robot may apply incorrect forces or fail to grasp an object, as it's trained on inaccurate collision models.
\end{marginnote}
}
{}
\subsection{Actuation and Control}
\label{subsec:actuation}
Actuators, together with the low-level control loops around them, are the last interface of the robot's actions and the real world.
They turn the actions from the policy into real-world motion and interaction.
Even when perception and dynamic models are perfect, any mismatch or gap between what the policy predicts will occur, and what actually happens at this interface can dominate the overall behavior of the robot and can lead to performance degradation.
Feasible commands in simulation can become delayed or unstable once they are filtered through real actuators.
These discrepancies constitute the \emph{actuation gap}.
\subsubsection{Actuator Models}
Most simulators treat motors as first-order systems that are able to perfectly produce torque responses instantaneously and linearly to the command signal.
However, real actuators (e.g., motors) behave as higher-order systems whose electrical and mechanical time constants introduce a non-negligible phase lag.
In addition, actuators are non-linear due to dead-zones, backlash, slew rate constraints, time constants, hysteresis, etc.
If we take into account physical mechanisms attached to the motor, such as gears, these problems only get worse, leading to added delays and increased steady state error.
\ifthenelse{\equal{\showmarginnote}{1}}{
\begin{marginnote}
    \footnotesize \Symptom
    The robot's movements may be jerky, delayed, or unstable, especially during high-speed maneuvers.
\end{marginnote}
}{}
\subsubsection{Low-level Control}
Control policies typically do not produce raw torque/force commands directly, and real robots rarely accept those commands directly.
Instead, there exist one or several low-level control layers that drive the actuator low-level control signal (e.g., PWM), and take as input a higher-level signal setpoint (e.g, position or velocity).
This conversion is generally done through dedicated hardware and firmware, for which most of the time access is restricted by the vendor.
Additionally, there are often hidden filters for anti-aliasing and resonance suppression, saturation, anti-windup logic, or protective non-linear effects like rate limiting.
\ifthenelse{\equal{\showmarginnote}{1}}{
\begin{marginnote}
    \footnotesize \Symptom
    A policy's commands may not produce the expected motion, or may even lead to instability, as they are modified by hidden filters on the real hardware.
\end{marginnote}
}{}
General robotics simulators do not simulate the lowest-level controllers, since they strongly depend on the manufacturer of the actuator itself.

\subsubsection{Power Electronics}
Between the low-level controller commands and the input signal to the motors, there is an additional layer: the power electronics.
Motor drivers, inverters, and Electronic Speed Controllers close their inner loops, introducing a latency of hundreds of microseconds.
Additionally, finite PWM resolution quantizes the commands produced by the actuator, reducing accuracy and introducing additional dead zones near zero crossings.
Finally, most drivers come with protection logic, which enforces hard current and voltage caps.
These effects are absent in simulations and can largely widen the actuation gap.
\ifthenelse{\equal{\showmarginnote}{1}}{
\begin{marginnote}
    \footnotesize \Symptom
    Fine-grained motions may fail, with jitter or dead zones near zero velocity.
\end{marginnote}
}{}

\subsection{System Design}
\label{subsec:system}
In addition to the gaps induced by the different modules of the robotics stack, the system design and the choices it entails can influence the reality gap.

\subsubsection{Communication}

Although communication in simulation is nearly perfect, communication in real-world robotic systems can face multiple challenges, including delays and packet loss.
In the real world, several mechanisms can be implemented to address these challenges.
For instance, it is common to introduce some kind of control attenuation mechanism to gradually bring the robot to a safe stop in case of prolonged communication packet loss.
Such mechanisms and behaviors are almost never modeled in simulation.
\ifthenelse{\equal{\showmarginnote}{1}}{
\begin{marginnote}
    \footnotesize \Symptom
    Communication delays may cause freezing, slowdown, or jerky fallback motions outside the trained policy.
\end{marginnote}
}{}

\subsubsection{Safety Mechanisms}

Real-world robotic experiments present several safety challenges that are not of concern in simulation.
It is very common to implement safety mechanisms such as virtual walls in the real world but not in simulation.
Such mechanisms change the behavior of the robot and can further widen the reality gap.
\ifthenelse{\equal{\showmarginnote}{1}}{
\begin{marginnote}
    \footnotesize \Symptom
    A policy may fail to react to safety mechanisms unseen during training in simulation.
\end{marginnote}
}{}

\subsubsection{POMDP Formulation}
Simulated POMDPs often feature unrealistic information and behaviors. 
For example, rewards and termination criteria may rely on privileged data, such as exact collisions, not accessible in the real world, leading to different rewards in simulation and reality. 
This mismatch is particularly problematic for model-based control methods that depend on reward signals at inference time. 
Similarly, environment resets may use infeasible behaviors, such as placing objects at precise positions with preset velocities, creating state-action distributions unlike those in real deployment, where such resets are impossible. 
\ifthenelse{\equal{\showmarginnote}{1}}{
\begin{marginnote}
    \footnotesize \Symptom
Real-world evaluations might strongly differ from the ones in simulation due to differences in the state-action distribution and the reward.
\end{marginnote}
}{}

\subsubsection{Implementation Details}
Many implementation details often differ between real-world and simulated environments. 
Simulation commonly involves the discretization of naturally continuous processes such as image formation and system dynamics.
The granularity of these discretizations can significantly impact the reality gap.
Low-level control implementation can also involve computational integration and differentiation steps, which can also end up running at different frequencies in simulation and the real world.
\ifthenelse{\equal{\showmarginnote}{1}}{
\begin{marginnote}
    \footnotesize \Symptom
A policy may become unstable due to mismatched control frequencies or numerical methods.
\end{marginnote}
}{}
\section{Existing Solutions} \label{sec:existing}

\begin{figure}
    \centering
    \includegraphics[width=\linewidth]{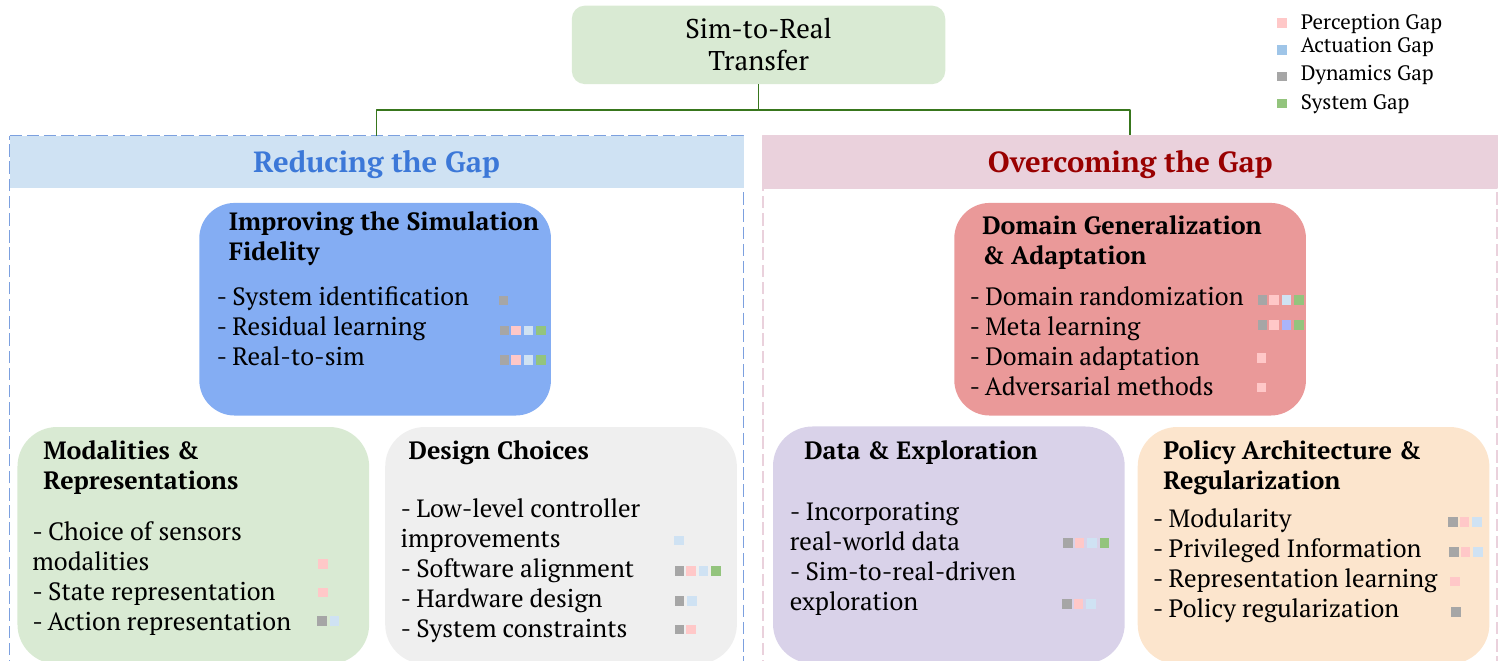}
    \caption{A taxonomy of sim-to-real transfer methods, distinguishing clearly between approaches and techniques that are designed to reduce the reality gap and those designed to overcome it.
    The colored squares indicate which reality gap each method addresses.}
    \label{fig:gap_minimize}
\end{figure}

Successful sim-to-real transfer requires careful consideration of the reality gap.
Methods addressing this challenge typically target one or more of the previously introduced sources of the reality gap (Section~\ref{sec:gaps}).
We mainly distinguish between approaches that attempt to \textit{reduce the gap} by improving different aspects of the simulation (i.e., eliminating sources of the reality gap) and approaches that attempt to \textit{overcome the gap} by making the system agnostic, adaptive, and/or passive to certain simulation aspects that cannot be accurately modeled.
Figure~\ref{fig:gap_minimize} summarizes the different methods that can be used for reducing and overcoming the different reality gaps.
A general recipe for successful sim-to-real transfer includes the following steps.

\begin{tcolorbox}
[boxrule=0pt, colback=shadecolor]
\textbf{Sim-to-Real Recipe}
\begin{enumerate}
    \item Design a simulation that represents all variables relevant for the target application.
    \item Attempt to reduce the different components of the reality gap as much as possible.
    \item Design training methods that can help overcome the remaining gap.
    \item Train policies in simulation (ideally under massive parallelization).
    \item Evaluate policies in the target real-world environment.
    \item Adjust simulation parameters based on real performance and retrain. 
\end{enumerate}
\end{tcolorbox}

\subsection{Reducing the Gap} \label{sec:reducing}

To reduce the gap, we need to make system modifications that either i) improve the simulation's fidelity (see section~\ref{sec:improvsim}), ii) select input and output modalities and representations that have a smaller reality gap (see section~\ref{sec:modandrep}), or iii) design systems with components and constraints that induce a smaller gap (see section~\ref{sec:designch}).

\subsubsection{Improving Simulation}
\label{sec:improvsim}
It is possible to improve different aspects of the simulated environment, such as its physical fidelity, sensor models, and many other components that influence the gaps introduced in section~\ref{sec:gaps}.

\vspace{0.5em}
\begin{itemize}[leftmargin=*, nosep]

\item \textbf{System Identification.} has been repeatedly shown to be a crucial aspect for successful sim-to-real transfer in multiple domains such as navigation~\cite{kaufmann2023champion}, locomotion~\cite{tan2018sim, hwangbo2019learning}, and manipulation~\cite{golemo2018sim,memmel2024asid}.\ifthenelse{\equal{\showmarginnote}{1}}{\begin{marginnote}[]
\textbf{Sys-ID} helps reduce the \textit{dynamics  gap}.
\end{marginnote}}{}One recent trend is to do online and iterative system identification with data collected at different iterations of real-world evaluations~\cite{farchy2013humanoid, chebotar2019closing, allevato2020iterative,du2021auto}.
In addition to physical parameters, it is common to identify and match system-level parameters representing different aspects of a physical environment, such as its latency, control frequency, and actuation delay~\cite{tan2018sim, hwangbo2019learning, lee2020learning,song2023reaching}.

\item 
\textbf{Learned Residual Models.}
When discrepancies between the simulated parametric model and the real-world are large, for example when the simulated physics model makes fundamentally-inaccurate assumptions (e.g., rigid-body dynamics for modeling compliant bodies), system identification may be ineffective.\ifthenelse{\equal{\showmarginnote}{1}}{\begin{marginnote}[]
\textbf{Learned residual models} can reduce several sources of \textit{reality gap}.
\end{marginnote}}{}Residual simulation, on the other hand, proposes a compelling solution: learn a model 
that modifies an imperfect simulator, such that the composite dynamics model more accurately reflects the real world.
Residual simulation approaches often use learning to modify the \textit{outputs} of a simulator, namely the predicted states, in order to better reflect real-world observations. 
Golemo et al.\ \cite{golemo2018neuralaugmented} first trained an Long-Short Term Memory neural network on differences between simulated trajectories and real-world trajectories to improve policy transfer.
Ajay et al.\ \cite{ajay2018augmenting} trained a stochastic Recurrent Neural Network to augment a deterministic simulator to capture uncertainty in contact dynamics. 
Gruenstein et al.\ \cite{gruenstein2021residual} trained a network to augment an analytical model 
to capture unmodeled compliance and friction. 
Bauersfeld et al.\ \cite{bauersfeld2021neurobem} trained a neural network to augment a blade-element-momentum aerodynamic model, capturing residual forces and torques for high-speed quadrotor flight.
Gao et al.\ \cite{gao2024softresidual} extended residual simulation to soft robotics, training a network to apply corrective forces to a finite-element model of a pneumatic arm.
Other residual simulation efforts have proposed a learning technique to modify the \textit{inputs} of a simulator, namely the applied actions, so that the resulting states more closely reflect the real world. For instance, 
Christiano et al.\ \cite{christiano2016transfer} trained policies for a manipulator in simulation, 
and afterwards learned an inverse dynamics model that maps a simulated action to a corrected real-world action that produces the state transition intended by the simulator.
Hanna et al.\ \cite{hanna2017gat} took a similar approach, but also augmented the simulator with the action transformation model before training the policy. 
\item \textbf{Real-to-Sim First.}
A variety of techniques have aimed to use real-world data to construct visually and dynamically accurate simulation environments on the fly at test time~\cite{torne2024reconciling, torne2024robotlearningsuperlinearscaling, dan2025xsimcrossembodimentlearningrealtosimtoreal, jiangphystwin, patel2025real, memmel2024asid, huang2023what, pfaff2025scalablereal2simphysicsawareasset, chen2022real2simsim2realroboticsvisual, chen2024urdformer, hsu2023ditto, zhao2024real2code}. 
Real-to-sim environment creation involves creating environments in simulation from data collected in the real world $\mathcal{D}_{\text{real}} = \{(z, a, z')_i\}_{i=1}^N$.\ifthenelse{\equal{\showmarginnote}{1}}{\begin{marginnote}[]
\textbf{Real-to-sim} can reduce various kinds of \textit{reality gap}.
\end{marginnote}}{}This ``inverse" problem has multiple facets -- constructing accurate geometry of the environment, constructing accurate kinematics and dynamics, and lastly constructing accurate visuals and rendering. 

For geometry identification, techniques for real-to-sim often build on 3D reconstruction~\cite{schoenberger2016sfm, schoenberger2016mvs} and novel-view synthesis methods for reconstructing geometry from multi-view data. 
Alternatively, a more recent class of monocular-3D methods has used learning-based priors to generate entire 3D meshes from single RGB images of objects~\cite{wang2018pixel2mesh}. 

In terms of identifying kinematics and dynamics, there is a range of techniques that have been proposed to take constructed geometries and imbue them with physical properties needed for manipulation. 
Kinematic identification involves accurately inferring object and scene articulation, and degrees of freedom from real data. 
This is particularly challenging when each of these DoFs is not directly activated during data collection ~\cite{hsu2023ditto}. 
A variety of recent methods ~\cite{chen2024urdformer, le2025articulate, zhao2024real2code} have attempted to infer articulation from video or image sequences using foundation models. While they provide a reasonable starting point, there is still a considerable gap with respect to the real world.
Finally, to obtain accurate environment renderings, novel-view synthesis methods have been particularly effective ~\cite{splatting, nerf}. 
These techniques enable high-fidelity neural rendering learned from purely real-world data, while enabling 3D, multiview image queries. 
\end{itemize}

\subsubsection{Choice of Modalities and Representations}
\label{sec:modandrep}
For most robotic tasks, it is possible to define different observation and action spaces.
This choice of interface defines the potential behaviors of the resulting system and can have strong implications on the reality gap~\cite{10522877,mahler2017dex,tai2017virtual,ding2021sim,kollar2022simnet,miki2022learning, agarwal2023legged,zhu2024point}.

\vspace{0.5em}
\begin{itemize}[leftmargin=*, nosep]
\item \textbf{Choice of Modalities.}
Different sensors and actuators can yield different gaps~\cite{zhou2019does,pmlr-v155-chen21f}.
For instance, using depth observationn, or point clouds yields a smaller reality gap than images, and this approach is a common practice in the field~\cite{loquercio2021learning, mahler2017dex, miki2022learning, agarwal2023legged, zhu2024point}.
RGB images are hard to render photo-realistically, and often include artifacts and features that are very complex to model, such as the ones discussed in section~\ref{sec:sensor_models}.
\ifthenelse{\equal{\showmarginnote}{1}}{
\begin{marginnote}[]
\textbf{Choice of modalities \& representations} affects the \textit{perception gap}.
\end{marginnote}
}{}
\item 
\textbf{State Representations.}
Similar to sensor modalities, the choice of state representations can strongly influence the reality gap~\cite{James_2019_CVPR,so2022simtoreal,zhangatk,ho2021retinagan,silwal2024we}. 
While using inferred depth maps can be a popular choice in the literature, other representations such as keypoint detections~\cite{zhangatk}, visual feature tracking~\cite{kaufmann2020RSS} or foundation model embeddings~\cite{silwal2024we} have also been proved to be good abstractions for reducing the sensory reality gap.

\item 
\textbf{Action Representations.}
The action space plays a crucial role in reducing the sim-to-real gap as demonstrated across robotics domains including navigation~\cite{kaufmann2022benchmark}, locomotion~\cite{kim2023torque}, and manipulation~\cite{10522877}.
For instance, for manipulation, previous studies have shown the advantages of using \ifthenelse{\equal{\showmarginnote}{1}}{\begin{marginnote}[]
\textbf{Choice of action representations} affects the \textit{actuation gap}.
\end{marginnote}}{}
joint velocity control spaces for sim-to-real transfer~\cite{10522877}.

\end{itemize}

\subsubsection{Design Choices}
\label{sec:designch}

Another important factor to reduce the reality gap is the system design and implementations.
When designing a robotic system, multiple choices need to be made depending on the task requirements.
This design, however, strongly influences the transferability of any behavior from simulation to the real world.
Hence, it is quite important to consider the reality gap when designing the system, including its hardware, the controller's implementations, constraints, and even the software stack.

\vspace{0.5em}
\begin{itemize}[leftmargin=*, nosep]
\item \textbf{Low-level Controller Improvements.}
While high-level policy learning typically takes place in simulation, the final interaction with the real world is governed by low-level controllers.
Improving the robustness and fidelity of these controllers is therefore crucial for overcoming the reality gap.
One effective approach is to increase the control frequency or bandwidth, allowing the system to respond more quickly to discrepancies and better handle latency, noise, or unmodeled dynamics.\ifthenelse{\equal{\showmarginnote}{1}}{\begin{marginnote}[]
\textbf{Low-level controller improvements} can reduce the \textit{actuation gap}.
\end{marginnote}}{}For example, Zhang et al.~\cite{zhang2024bridging} show that high-frequency impedance controllers can significantly improve sim-to-real transfer in manipulation tasks by stabilizing behavior under real-world disturbances.
Such improvements on the low-level controllers can make the system more resilient to imperfections in both the simulator and the real-world environment.
\item \textbf{Software Stack Alignment.}
Beyond physical realism, achieving successful sim-to-real transfer also requires consistency in the software stack between simulation and real-world deployment.\ifthenelse{\equal{\showmarginnote}{1}}{\begin{marginnote}[]
\textbf{Software stack alignment} can reduce both \textit{reality gap} and \textit{performance gap}.
\end{marginnote}}{}Even small discrepancies, such as mismatched control rates, missing filters, or unmodeled safety checks, can lead to unexpected behavior and degraded performance.
Replicating consistent software components ensures that the policy is exposed to the same control dynamics during both training and real-world execution.
\item
\textbf{Hardware Design.}
The design of the robot itself can also significantly influence the sim-to-real gap.
Hardware choices that simplify modeling or improve physical robustness can make it easier to simulate the system accurately and improve robustness to discrepancies.\ifthenelse{\equal{\showmarginnote}{1}}{\begin{marginnote}[]
\textbf{Hardware design} can affect the of \textit{dynamics gaps} and the \textit{actuation gap}.
\end{marginnote}}{}For example, using actuators with low latency and consistent torque output, or sensors with well-characterized noise profiles, can improve the fidelity of simulation and reduce the need for complex system identification.
Additionally, hardware with passive stability, compliant actuators, or simple kinematics can better tolerate control and perception errors, which could further improve the chances of successful transfer.
These principles align with recent advances in real-world imitation learning, such as~\cite{chi2024universal}, where careful co-design of hardware and learning pipelines enables effective policy learning.
\item
\textbf{Constraining System Dynamics.}
Another practical strategy for reducing the reality gap is to initially limit the robot’s dynamic complexity during deployment.
By operating at lower speeds or avoiding aggressive maneuvers, the system becomes less sensitive to modeling inaccuracies, actuator delays, and perception noise.
Such non-dynamic behaviors reduce the burden on both the controller and simulator, improving robustness under real-world uncertainties.\ifthenelse{\equal{\showmarginnote}{1}}{\begin{marginnote}[]
\textbf{Constraining  system dynamics} can reduce the \textit{dynamics gap} and the \textit{perception gap}.
\end{marginnote}}{}For example, Chen et al.~\cite{chen2021hardware} show that constraining motion to quasi-static regimes can lead to more reliable sim-to-real transfer in manipulation tasks.
\end{itemize}

\subsection{Overcoming the Gap}

In the previous section, we discussed different paradigms to reduce the reality gap with different measures to bring the simulated and real-world environments closer together.
An orthogonal family of methods assumes a fixed reality gap and attempts to overcome it by either making the policies agnostic to the choice of models and parameters or making the policy capable of detecting and reacting to different physical and systematic parameters.

\subsubsection{Domain Generalization and Adaptation}

One of the most common approaches to overcome the reality gap is to expose the policy to a large variation of the system and dynamics parameters during training.
As a result, the policy should be robust to different instantiations of these parameters, including the ones representing the real world.
\vspace{0.5em}
\begin{itemize}[leftmargin=*, nosep]
\item
\textbf{Domain randomization~(DR).}
Among the most popular and widely-used approaches to achieve this is DR.
In DR, we train the policy in simulation using a broad distribution of simulated environments spanning different simulation parameters such as visual (more generally sensory) parameters like 
texture and lighting, observation and action noise, physics parameters such as mass and\ifthenelse{\equal{\showmarginnote}{1}}{\begin{marginnote}[]
\textbf{Domain randomization} can reduce all kinds of \textit{reality gaps}.
\end{marginnote}}{}inertia, system delays, object properties, and others.
Consequently, the policy can learn a behavior that is applicable to such a large distribution of environments.
DR mitigates the need for painstaking system identification and simulator calibration~\cite{peng2018sim}.
One of the earliest applications of DR was to drone navigation~\cite{sadeghi2017cad2rl}.
The paradigm has gained significant popularity ever since and has been used to produce some of the most impressive robotics milestones, including dexterous manipulation~\cite{akkaya2019solving}, quadruped locomotion~\cite{tan2018sim, hwangbo2019learning}, and champion-level drone racing~\cite{kaufmann2023champion}.
To further improve traditional DR, multiple approaches have been proposed to automate the choice of parameter distributions.
For instance, Akkaya et al.~\cite{akkaya2019solving} proposed automatic DR to progressively expand the range of random dynamics as the policy becomes successful.
Tiboni et al.~\cite{tiboni2023dropo} propose to use an offline real-world dataset to estimate the optimal DR ranges allowing to compensate for unmodeled effects.
Similarly, Chebotar et al.~\cite{chebotar2019closing} proposed using real-world data to infer simulation parameters for online training.
The lowest-return dynamics from an ensemble effectively acts like adversarial samples for training. For a comprehensive review of DR methods, see~\cite{muratore2022robot}.
\item 
\textbf{Adversarial Training} systematically generates data to enforce robustness and enhance model resilience beyond standard training.
Pinto et al.~\cite{pmlr-v70-pinto17a} propose robust adversarial reinforcement learning, where in addition to the main agent, an adversary agent is trained to apply disturbances to destabilize the main agent's policy.\ifthenelse{\equal{\showmarginnote}{1}}{\begin{marginnote}[]
\textbf{Adversarial training} can reduce the \textit{actuation gap}, \textit{perception gap}, and \textit{dynamics gap}.
\end{marginnote}
}{}
\item \textbf{Meta Learning} is another paradigm for domain adaptation.
For instance, in~\cite{arndt2020meta}, the authors propose to use meta-learning for policy learning in simulation under a large distribution of simulation parameters, allowing the policy to internally recognize variations of parameters and act according to its internal understanding.\ifthenelse{\equal{\showmarginnote}{1}}{\begin{marginnote}[]
\textbf{Meta learning} adapts the simulation parameters to reduce the \textit{reality gap}.
\end{marginnote}}{}This family of methods leverages meta reinforcement learning to adapt the policy to variations of simulation parameters~\cite{renadaptsim}.
While domain-randomization-based methods aim to learn a policy that can generalize to a large distribution of simulation parameters, methods such as meta-learning aim to adapt the policy as inspired by adaptive control methods~\cite{aastrom1995adaptive,40741}. 
One famous example of such methods beyond explicit meta learning, is rapid motor adaptation~(RMA)~\cite{kumar2021rma}. RMA learns an encoder that infers latent representations of the environment's dynamics using privileged information in simulation. 
Given such an encoder, the policy can adapt its actions based on the inferred latent environment representations.

\item 
\textbf{Domain Adaptation} can similarly be used to enhance the adaptiveness of the policy to variations in environment variables and dynamics~\cite{rusu2017sim,bousmalis2018using,chebotar2019closing,zhang2019adversarial}.
Compared to meta learning and RMA, domain adaptation methods for sim-to-real transfer aim to make the policy more robust to the sim-to-real distribution shift.
In other words, the main focus of these methods\ifthenelse{\equal{\showmarginnote}{1}}{\begin{marginnote}[]
\textbf{Domain adaptation} can reduce the \textit{perception gap}.
\end{marginnote}}{} is to enable observation adaptation in the policy and not necessarily adaptation to the changing dynamics.

\end{itemize}
\subsubsection{Data Selection and Exploration}
Another important category of methods to overcome the reality gap is to carefully select and curate the training data and exploration strategies used in simulation.
These methods focus on generating training data that most closely resemble the target real-world data or data from worst-case behaviors.

\vspace{0.5em}
\begin{itemize}[leftmargin=*, nosep]
\item 
\textbf{Incorporating real-world data} to inform simulation training can greatly improve transfer.
For instance, Niu et al.~\cite{niu2022trust} proposed integrating limited real-world data with simulated experiences, while adaptively penalizing learning from simulated state-action pairs that exhibit significant discrepancies from real-world dynamics.
Torne et al.~\cite{torne2024reconciling} proposed a real-to-sim-to-real pipeline that first uses real-world data to train a vision-based policy that is later used to train a \ifthenelse{\equal{\showmarginnote}{1}}{\begin{marginnote}[]
\textbf{Real world data} can be used to reduce all kinds of \textit{reality gaps}.
\end{marginnote}}{}teacher policy in simulation. The teacher is then used to train a vision-based student policy in simulation using a mixture of RL with simulated rollouts and behavior cloning using real-world data.
This work demonstrated the benefits of co-training with real-world data versus training with only simulation data.
Ankile et al.~\cite{ankile2024imitation} proposed a residual RL scheme to combine real-world and simulated data for vision-based manipulation.
Maddukuri et al.~\cite{maddukuri2025sim} demonstrated the effectiveness of co-training vision-based manipulation policies with simulation and real-world data.
Co-training has also been shown to be beneficial for training robotic foundation models~\cite{robocasa2024,gr00t2025}.

\item
\textbf{Sim-to-real-driven exploration} methods aim to explore the state-action space in ways that expose the policy to interactions likely to enhance its transfer from simulation to the real world.
For instance, Liang et al.~\cite{liang2020learning} proposed learning exploration policies that are executed in the real world to identify critical system parameters.
Given this improved model of the environment, they perform trajectory optimization to solve the downstream tasks in the real-world environment.\ifthenelse{\equal{\showmarginnote}{1}}{\begin{marginnote}[]
\textbf{Sim-to-real-driven exploration} can reduce the \textit{perception gap}, \textit{dynamics gap}, and \textit{actuation gap}.
\end{marginnote}}{}Another approach is to leverage the simulation to learn exploration policies that are transferred to the real world for the sole purpose of learning and fine-tuning policies with real-world interactions~\cite{wagenmaker2024overcoming,memmel2024asid}.

\end{itemize}

\subsubsection{Policy Architecture and Regularization}

In addition to the data a policy is exposed to at training time, the policy's architecture and constraints can play a crucial role in the transfer to the real world.
In this section, we describe different ways to structure the policy and regularize it in a way that helps sim-to-real transfer.

\vspace{0.5em}
\begin{itemize}[leftmargin=*, nosep]
\item 
\textbf{Modularity} of the system and policy architecture has also been shown to be beneficial for sim-to-real transfer.
Clavera et al.~\cite{clavera2017policy} proposed decomposing the system into distinct modules, where vision and low-level control are managed independently from the policy.
Their work demonstrated robust transfer of pushing policies.
Mueller et al.~\cite{mueller2018driving} proposed a similar architecture for autonomous driving.\ifthenelse{\equal{\showmarginnote}{1}}{\begin{marginnote}[]
\textbf{Modularity} can reduce the \textit{perception gap}, \textit{dynamics gap}, and \textit{actuation gap}.
\end{marginnote}}{}Similarly, Zhang et al.~\cite{zhang2017modular} propose learning perception and control networks with different losses and fine-tuning the end-to-end policy network combining perception and control modules using a weighted loss.
Julian et al.~\cite{julian2020scaling} proposed a hierarchical architecture, learning several low-level skills and a high-level policy coordinating them.
By decomposing the policy into multiple skills, they reduce the amount of data needed to generalize to a new domain and environment, such as the real world.

\item
\textbf{Privileged information} is available to the agent at training time in simulation but not during deployment in the real world.
Such information can boost learning efficiency and enable larger-scale training in simulation~\cite{chen2020learning, hu2024privileged,messikommer2025studentinformed}.
One popular example of such method is the previously discussed RMA algorithm~\cite{kumar2021rma}.
RMA uses privileged information to train a latent space online as part of a system identification module. 
Pinto et al.~\cite{pinto2018asymmetric} propose an asymmetric actor-critic architecture, where the critic is conditioned on privileged information, while the actor policy is\ifthenelse{\equal{\showmarginnote}{1}}{\begin{marginnote}[]
\textbf{Privileged information} addresses the \textit{perception gap}, \textit{dynamics gap}, and \textit{actuation gap}.
\end{marginnote}}{}conditioned on true observations.
Radosavovic et al.~\cite{radosavovic2024real} trained a teacher policy for humanoid locomotion with privileged information and then distilled the teacher behavior into a student agent conditioned on the observations.
To overcome student-teacher asymmetry, they propose a loss combining RL and teacher distillation, similar to the approach previously proposed in~\cite{nguyenleveraging} for block picking.
The main difference between these two methods is how they combine RL and imitation learning losses using either fixed coefficients~\cite{nguyenleveraging} or a schedule~\cite{radosavovic2024real}.
More recently, Krinner et al.~\cite{krinner2025accelerating} proposed using privileged state information in state-space world models to improve the training efficiency of a model-based RL agent.
The proposed approach was successfully demonstrated on vision-based high-speed drone racing.
\item
\textbf{Representation learning} leverages self-supervised objectives to learn representations that can be better suited for policy search and sim-to-real transfer.
By carefully designing the representation learning loss, we can enforce the learned representation properties that are desirable to overcome certain sources of the reality gap.
While domain randomization and adaptation methods improve domain robustness via exposing the policy to a wider set of data with the hope of out-of-distribution generalization, representation learning methods leverage loss functions to learn robust features.\ifthenelse{\equal{\showmarginnote}{1}}{\begin{marginnote}[]
\textbf{Representation learning} can help reduce the \textit{perception gap}.
\end{marginnote}}{}For instance, Tanwani et al.~\cite{tanwani2021dirl} proposed an approach that leverages real-world data to align the distributions of the state representations in simulation and the real world, making the features more robust to the corresponding domain shift.
Yoneda et al.~\cite{yoneda2022invariance} propose to use adversarial training to adapt the observation encoder learned in simulation once it encounters real-world samples.
This process is combined with a dynamics consistency loss to ensure that latent transitions in the target domain remain faithful to the dynamics learned in the source domain, thereby preserving policy effectiveness despite visual discrepancies.
Xing et al.~\cite{xing2024contrastive} proposed a contrastive learning approach to learn background-agnostic and task-relevant representations, ensuring effective feature learning for the downstream task while ignoring irrelevant, noisy background information.

\item
\textbf{Policy regularization} can change the behavior of the policy in a manner that helps overcome the reality gap for sim-to-real transfer.
Regularization typically occurs through loss functions or elaborate reward designs for RL agents.
For instance, it has become a common practice to penalize the magnitude of actions and consecutive action differences through reward terms or penalties~\cite{hwangbo2019learning,song2023reaching,xing2024multi,handa2023dextreme,10522877}.\ifthenelse{\equal{\showmarginnote}{1}}{\begin{marginnote}[]
\textbf{Policy regularization} can reduce the \textit{dynamics gap}.
\end{marginnote}}{}Some works even penalize the magnitudes of explicit measures such as velocity and accelerations~\cite{handa2023dextreme,hwangbo2019learning}, as well as power loss~\cite{aractingi2023controlling}.
While all of these works rely on RL penalties to enforce desirable properties such as smoothness on the learned policies, such policies can also be imposed using loss functions.
For instance, Mysore et al.~\cite{mysore2021regularizing} introduced smoothness losses to the policy training algorithm to minimize the temporal and spatial 
Lipschitz constraints of a quadrotor control policy function.
Their work demonstrated a clear advantage of loss function regularization over reward engineering.
Chen et al.~\cite{chen2024lcp} presented similar findings for humanoids locomotion.
While these methods enforce properties on the policy to overcome the reality gap, Niu et al.~\cite{niu2022trust} proposed incorporating real-world data during training to penalize the action-value function in simulation when interactions exhibit significant discrepancies in dynamics between the simulated and real environments.

\end{itemize}
\section{Evaluation  Metrics}
\label{sec:evaluation}
Research on sim-to-real transfer has leveraged various metrics for evaluation.
In this section, we survey such metrics while making a clear distinction between metrics evaluating the reality gap itself and metrics that evaluate the sim-to-real transfer performance.

\subsection{Assessing the Reality Gap}
A well-designed evaluation framework for assessing the gap between simulation and the real world is critical for reducing the reality gap, as well as for understanding, diagnosing, and ultimately improving the transferability of learned policies.
\vspace{0.5em}
\begin{itemize}[leftmargin=*, nosep]
\item
\textbf{Sim-to-real Correlation Coefficient.}
One key question in sim-to-real transfer is whether performance improvement observed in simulation reliably leads to better performance in the real world.
To address this, Kadian, et al.~\cite{kadian2020sim2real} introduced the sim-to-real Correlation Coefficient (SRCC).
SRCC is defined as the Pearson correlation coefficient between the performance metrics of the agents in simulation and the real world,
\begin{equation}
\text{SRCC} = \frac{\sum_{i=1}^{N} (x_i - \bar{x})(y_i - \bar{y})}{\sqrt{\sum_{i=1}^{N} (x_i - \bar{x})^2} \sqrt{\sum_{i=1}^{N} (y_i - \bar{y})^2}},
\end{equation}
where \(x_i\) and \(y_i\) denote the the task evaluation performance (e.g., success rates) of the \(i\)-th policy in simulation and the real world, respectively, and \(\bar{x}\), \(\bar{y}\) are their corresponding means.
A coefficient close to $+1$ indicates a strong positive correlation, suggesting that the simulation performance is a good predictor of real-world performance.
In contrast, values near $0$ imply poor correlation, meaning simulation results offer little guidance for real-world evaluation.
Importantly, a simulator with high average real-world performance but low SRCC is still problematic.
Without a reliable correlation, improvements in simulation may have unpredictable effects in the real world, making it difficult to make informed design decisions.
Most of the changes need to be validated on a physical robot, which defeats the purpose of using simulation to accelerate development.

\item
\textbf{Offline Replay Error.}
When direct real-world policy deployment is not feasible, offline replay error provides a practical alternative for evaluating sim-to-real transfer. 
This situation often arises during early prototyping stages or when deployment is limited by cost, safety concerns, or hardware availability.
Offline replay error compares state trajectories between a real-world trajectory and the equivalent simulation trajectory when replaying the real-world actions in an open-loop fashion~\cite{10522877}. 
Formally, given a trajectory of real-world policy rollout states \(\{\mathbf{s}_t^{\text{real}}\}\) and corresponding actions \(\{a_t\}\), the offline replay error is defined as
\begin{equation}
\mathcal{E}_{\text{replay}} = \frac{1}{T} \sum_{t=1}^{T} \| \mathbf{s}_t^{\text{sim}} - \mathbf{s}_t^{\text{real}} \|^2,
\end{equation}
where \(T\) is the length of the trajectory, and $\mathbf{s}_t^{\text{sim}}$ is obtained by rolling out the real-world actions in an open-loop fashion in simulation.
This metric is appealing due to its simplicity and low cost: no real-time interaction is required, and evaluation can be done offline using logged data.
It provides a quick diagnostic of how well the simulation represents the target real-world environment under the distribution induced by the policy. 
A policy having a high offline replay error is a strong indicator of a large reality gap.
\item
\textbf{Visual Fidelity Analysis.}
Assessing the perceptual domain gap between simulated and real-world environments is particularly relevant for visuomotor policies that map visual observations directly to control commands, as well as visual state representations methods.
To quantify the visual similarity, a variety of metrics have been proposed that operate at either the pixel level or in the space of learned image embeddings.
These include both distribution-level metrics, which assess the global characteristics of image sets, and single-image metrics, which compare individual simulated images to corresponding real ones~\cite{lambertenghi2024assessing}.
For distribution-level evaluation, commonly used metrics include Inception Score (IS)~\cite{salimans2016improved}, Fréchet Inception Distance (FID)~\cite{heusel2017gans}, and Kernel Inception Distance (KID)~\cite{yu2020toward}, TSNE dimensionality reduction~\cite{van2008visualizing, biruduganti2025bridging, xing2024contrastive}, all of which operate on feature representations extracted from pretrained networks.
For single-image evaluation, metrics such as the Structural Similarity Index (SSIM) and Peak Signal-to-Noise Ratio (PSNR)~\cite{salimans2016improved}, Instance Performance Difference (IPD)~\cite{chen2024instance} are used to measure differences in luminance, contrast, and structural details.
\end{itemize}
\subsection{Assessing Sim-to-Real Transfer}
While evaluating the reality gap is very valuable to analyze the limitations of the simulation and further improve it, metrics that evaluate the sim-to-real transfer performance under a fixed reality gap are also necessary.
This category includes metrics such as success rate, cumulative reward (in reinforcement learning), and task-specific indicators tailored to different domains.
In this section, we will discuss in detail the evaluation metrics commonly used to quantify sim-to-real transfer performance.
\vspace{0.5em}
\begin{itemize}[leftmargin=*, nosep]
\item
\textbf{Success rate}
 measures the proportion of trials in which a policy successfully completes the intended task, written as
$
\text{Success Rate} = \frac{N_{\text{success}}}{N_{\text{total}}}.
$
It is widely used as a simple and reliable indicator of sim-to-real transfer effectiveness.
This measure is most informative when proper randomization is performed on all involved variables, such as the initial environment state and simulation seeds, among many others.
Success rates are commonly reported in diverse domains, including manipulation~\cite{james2019sim, fu2023deep, alles2022learning, andrychowicz2020learning, ding2021sim, lin2025sim, dalal2024local, singh2024dextrah}, navigation~\cite{zeng2024poliformer, gervet2023navigating,kadian2020sim2real, chang2024goat}, autonomous racing~\cite{kaufmann2023champion, xing2024bootstrapping}, legged locomotion~\cite{kumar2021rma, hwangbo2019learning, kumar2021rma}.
High task success rates in the real world often indicate that a policy has successfully transferred key competencies learned in simulation.
However, such metrics are often binary or aggregate and do not reveal where or why the transfer might have failed.
For example, two policies with the same success rate might differ significantly in robustness, and some of the failure modes can be critical in real-world deployment.
\item
\textbf{Cumulative reward.}
For policies learned using \emph{reinforcement learning (RL)} or with a predefined reward structure, the cumulative reward measures how well a policy achieves long-term objectives by summing the reward signal over time:
$
R = \sum_{t=0}^{T} r_t,
$
where 
$r_t$ is the reward at timestep 
$t$ and 
$T$ is the total duration of the episode.
Unlike binary success metrics, it can capture the progress and efficiency of the policy rollouts, offering a more fine-grained view of sim-to-real performance.
However, its interpretability depends on consistent and well-designed reward functions across simulation and real-world domains~\cite{sandha2021sim2real, wagenmaker2024overcoming, chebotar2019closing}. 
This can make the comparison of different sim-to-real techniques under varying RL configurations difficult, especially when reward formulations differ across tasks or platforms.
\item
\textbf{Task-specific metrics.}
In addition to general success or reward-based metrics, most of the successful sim-to-real studies evaluate performance using task-specific criteria tailored to the target domain.
For example, navigation tasks may use path efficiency or time-to-goal~\cite{gervet2023navigating}, while manipulation tasks typically use object-centric metrics, such as object-distance-to-goal for object pushing~\cite{10522877}.
These metrics offer fine-grained insights into policy behavior and failure modes that are not captured by success rate alone.
However, for real-world deployment, it will be challenging to provide standardization across benchmarks, which can hinder fair comparison between methods.
\end{itemize}

\section{Discussion and Open Problems} \label{sec:open}
Despite the reality gap and the challenges it creates, sim-to-real transfer has been a very successful and popular paradigm in robotics.
However, it is unclear whether simulation will remain a major tool for robotics development as many challenges persist. We believe that there is still untapped potential for simulation in robotics and discuss some of the challenges and opportunities next.
\subsection{Wrong Models, Better Controllers}
As briefly mentioned in section~\ref{sec:realitygap}, it is more important in practice to reduce the sim-to-real performance gap than to reduce the reality gap. 
This motivates the question of whether accurate physics modeling is needed to perform reliable model-based control. 
Model-based approaches, such as model-based RL and model predictive control~(MPC) tend to be more data efficient than model-free RL and thus leveraging them in sim-to-real contexts can decrease the scale of data generation necessary for sufficiently reliable control.
Particularly, researchers have examined how training for model-based RL can be adapted to not focus on accurately capturing the underlying physics, but instead learning a dynamics model that improves control performance~\cite{lambertL4DC2020,guzmanICRA2022}. 
Results show that focusing on learning a model accurately near high-return areas outperforms the alternative of trying to learn a model with uniform accuracy across the state-action space~\cite{lambertL4DC2020}. 
Guzman et al.~\cite{guzmanICRA2022} examine the use of Bayesian optimization for MPC, where they sample different physical parameters in simulation for tuning the MPC controller. 
Results show that optimizing for control performance outperforms planning with the most likely model estimate. 
As such, an open question remains as to how a robot can most efficiently use a simulator with incorrect parametrization to learn either a policy or stochastic world model for use in generating robust, real-world performance.

\subsection{Differentiable Simulators.}
Differentiable simulators offer the computation of gradients of quantities involved in the simulation which allows them to be integrated in gradient-based optimization workflows~\cite{newbury2024diffsimreview}.
While traditional simulators, such as MuJoCo~\cite{todorov2012mujoco} and Bullet~\cite{coumans2021} can often be useful in such applications by leveraging finite differences, a \emph{differentiable} simulator typically offers fast and analytically correct gradients. 
The methods to compute such derivatives typically fall into two categories: automatic differentiation leveraging the chain rule of differentiation, and manual computed analytical gradients.
The most common form of Autodiff, reverse-mode automatic differentiation, accumulates gradients for the inputs from each operation, given the output gradients. 
Autodiff can be realized through a tape to record the operations of the computation path which allows the (reverse) playback in the backward pass. 
Frameworks, such as NVIDIA Warp~\cite{nvidia_warp}, Taichi~\cite{hu2019difftaichi}, 
JAX~\cite{jax2018github}, and PyTorch's compilation option~\cite{torchdynamo2023} allow code to be generated for the backward pass, which is critical for performance.
As such, an open question remains as to how to best leverage differentiable simulation and augment it with learning-based dynamics model.

\subsection{Video and World Models}
Video models are primarily designed to process, generate, or predict sequences of visual frames. 
Their core objective is to model temporal dynamics and evolving visual content within video data. 
These models address a wide range of tasks, including future frame prediction based on observed sequences and conditional video generation from text, actions, or other modalities~\cite{ming2024survey}. World models pursue a broader goal: learning internal representations of an environment that enable simulation and prediction of future states in response to agent actions or external events~\cite{bruce2024genie,hassan2025gem, agarwal2025cosmos}. 
These models seek to encode the causal structure of the world to support planning, imagination, and counterfactual reasoning~\cite{ding2024understanding}. %
Despite recent progress, key challenges remain with significant opportunities for future research. 
Video models, while effective at generating dynamic visual content, struggle with maintaining temporal consistency, physical plausibility, fine-grained controllability, and computational efficiency—issues that hinder their reliability in robotics and simulation. 
World models offer a promising alternative to white-box simulators and can be learned using real-world data, potentially creating a smaller reality gap.
However, they face problems such as compounding errors in long-term predictions, poor generalization to new environments, difficulty balancing abstraction with realism, and high data requirements. Addressing these challenges will require innovations in model architecture, integration of physical priors, and improved training strategies. In future iterations, the paradigms of world modeling and simulation may not be so distinct, with simulators naturally providing data to bootstrap world models for deployment in reality. 

\subsection{Simulation-Based Inference}
Simulation-based inference~\cite{cranmer_frontier_2020} is a statistical technique to approximate 
the posterior distribution of simulation parameters $\theta$. Given
a \emph{prior} distribution $p(\theta)$, simulation dynamics model $\mathcal{T}_\text{s}(s_{t+1}|s_t,a_t)$, a set of simulation observations $\{\mathbf{s}^\text{s}_i\}_{i=1}^{S}$ and a set of real state observations $\{\mathbf{s}^\text{r}_i\}_{i=1}^{R}$, the posterior can be computed following the Bayes' rule as
$p(\theta|\{\mathbf{s}^\text{r}_i\}_{i=1}^\text{R}, \{\mathbf{s}^\text{s}_j\}_{j=1}^\text{S}) \propto p(\{\mathbf{s}^\text{r}_i\}_{i=1}^\text{R}|\theta)p(\theta).$
The main challenge is that the likelihood term is generally intractable. 
If we consider simulation as a generative process, $\mathbf{s}^\text{s}=g(P_\text{s}, \theta)$, where $\mathbf{s}^\text{s}$ are simulated states or observations, the likelihood term can be defined as $p(\{\mathbf{s}^\text{r}_i\}_{i=1}^\text{R}|\theta)=\int p(\{\mathbf{s}^\text{r}_i\}_{i=1}^\text{R}, \mathbf{s}^\text{s}|\theta)d\mathbf{s}^\text{s}$. To compute this explicitly would require integrating over all possible trajectories the simulator could generate for a specific parametrization $\theta$.
This is not feasible for the majority of robotics simulators that involve complex physics models (including contact models), observation models, and numerical solvers. 
To address this issue, approximate inference methods have been proposed based on Monte Carlo techniques such as Approximate Bayesian Computation~\cite{sisson2018handbook}, and variational inference~\cite{glockler2022variational}. 
Also, note that computing the posterior is an ill-posed inverse problem, as there are possibly many simulation parameters that can generate the same trajectories. 
For example, in a pushing task where a manipulator pushes a block on a table, the resistance of the motion can be attributed to higher mass, higher friction, or both. 
This leads to posterior distributions that are highly multimodal. 
To account for generic posteriors and sidestep the approximation of the likelihood term directly, recent works in robotics have used techniques where a neural network is trained to directly output the posterior distribution $p(\theta|\{\mathbf{s}^\text{r}_i\}_{i=1}^\text{R}, \{\mathbf{s}^\text{s}_j\}_{j=1}^\text{S})$. 
This is known as neural posterior estimation~\cite{papamakarios2016fast, radev2022bayesflow} and has been used in robotics recently~\cite{ramos2019_bayessim, barcelos2020disco, matl2020inferring, possas2020online, antonova2021bayesian, muratore22a-neuralposterior, matl2021stressd}. 
The posterior can then be used as the randomization distribution for domain randomization. 
For a review of simulation-based inference methods applied to domain randomization, see~\cite{muratore2022robot}. 
We believe simulation-based inference methods will continue to be a popular research topic in robotics. 

\subsection{Simulation for Large Robotics Models}

In recent years, there have been substantial efforts to collect real-world data to train large models for robotics with techniques such as imitation learning~\cite{o2024open,khazatsky2024droid,hassan2025gem}.
However, realistically, real-world data collection of action-labeled datasets is limited by several factors such as human efforts and hardware resources.
Simulation represents a great opportunity to augment such efforts with massive-scale synthetic data generation.
Many recent works explored this avenue and proposed novel pipelines to procedurally generate simulated data based on real-world demonstrations~\cite{liudexscale,mandlekar2023mimicgen}.
As with RL, simulated data would need to have a small reality gap to be valuable in real-world deployments.
An open question is whether the methods and assumptions needed to reduce these gaps would themselves be limiting factors to the scale of data that can be collected in simulation (under these assumptions and using these methods).

Another recent trend is to leverage simulation to evaluate real-world policies in a systematic and reproducible manner~\cite{li24simpler,barreiros2025careful}.
In this context, a proper calibration of the reality gap is crucial for simulation to be a good proxy for the real world. 
While methods to reduce this gap are very similar to the ones used for sim-to-real transfer, an open question is how to overcome the reality gap for the specific downstream purpose of model evaluations such that the performance of a policy in simulation matches the performance in the real environment.

\section{Conclusion}
This survey is an attempt at dissecting the very complex yet very important problem of the reality gap in robotics.
We discussed the sources of these gaps, the problems they create and solutions to alleviate them, together with evaluation metrics and opportunities for future research.
By better understanding the problem, we can more clearly situate ourselves as a community to face future research challenges. Leveraging simulation to the best of our capabilities will enable a cost-effective alternative for the development of the future generation of robotics systems.

\section*{DISCLOSURE STATEMENT}
The authors are not aware of any affiliations, memberships, funding, or financial holdings that might be perceived as affecting the objectivity of this review.


\section*{AUTHOR CONTRIBUTIONS}
EA and FR defined the structure of the survey and its focus.
EA wrote the abstract, sections 1, 4, and 7 and contributed subsections to sections 2, 3, and 6. He also revised and updated many paragraphs throughout the text.
EA and JX made the illustrations.
JX wrote sections 3.2, section 5, and many paragraphs throughout the text.
AR wrote most of section 2, section 3.1, and section 3.3.
IA wrote section 6.3.  
CG provided general comments and suggestions and updated section 1. 
EH wrote section 6.2. 
AG contributed to different parts of section 4. 
TH wrote section 6.1. YN contributed to section 4.1.
DF and DS provided feedback, corrections and suggestions on the content of the article. 
FR conceptualized the work with EA, wrote section 6.4, and revised and updated several parts of the article.
\section*{ACKNOWLEDGMENTS}
This work was supported by the European Research Council (ERC) under grant
agreement No. 864042 (AGILEFLIGHT).



%
\bibliographystyle{ar-style3}
\bibliography{references}

\end{document}